\documentclass[lettersize,journal]{IEEEtran}
\usepackage{amsmath,amsfonts}
\usepackage{algorithmic}
\usepackage{algorithm}
\usepackage{array}
\usepackage{textcomp}
\usepackage{stfloats}
\usepackage{url}
\usepackage{verbatim}
\usepackage{graphicx}
\usepackage{subfigure}
\usepackage{multirow}
\usepackage{multicol}
\usepackage{booktabs}
\usepackage{color}
\usepackage{cite}
\usepackage{amsmath}
\usepackage{bbm}
\usepackage{enumerate}
\usepackage{enumitem}
\usepackage{amsfonts}
\usepackage{color}

\begin{document}

\title{Towards Effective Image Manipulation Detection with Proposal Contrastive Learning}

\author{Yuyuan Zeng$^{\star}$, Bowen Zhao$^{\star}$, Shanzhao Qiu, Tao Dai, Shu-Tao Xia.

\thanks{
Yuyuan Zeng and Bowen Zhao are equal contributions. 
This work is supported in part by the National Natural Science Foundation of China under Grant 62171248, Shenzhen Science and Technology Program (Grant No. JCYJ20220818101012025) and the PCNL KEY project (PCL2021A07).
 (\emph{Corresponding author: Tao Dai}) 
		\par
			Y. Zeng, and T. Dai are with the College of Computer Science and Software Engineering, Shenzhen University, Shenzhen 518060, China. Work was done when Y. Zeng was a master student at Tsinghua University.(e-mail: yuyuanzeng17@gmail.com; daitao.edu@gmail.com).
	\par
			B. Zhao, S. Qiu, S. Xia are with the Tsinghua Shenzhen
International Graduate School, Tsinghua University, Shenzhen
518055, China. S. Xia  is also with Research Center of Artificial Intelligence, Peng Cheng
Laboratory, Shenzhen 518055, China.
(e-mail: zbw18@mails.tsinghua.edu.cn; xiast@sz.tsinghua.edu.cn).
}
}
\markboth{Journal of \LaTeX\ Class Files,~Vol.~14, No.~8, August~2021}%
{Shell \MakeLowercase{\textit{et al.}}: A Sample Article Using IEEEtran.cls for IEEE Journals}


\maketitle

\begin{abstract}
Deep models have been widely and successfully used in image manipulation detection, which aims to classify tampered images and localize tampered regions. Most existing methods mainly focus on extracting \textit{global features} from tampered images, while neglecting the \textit{relationships of local features} between tampered and authentic regions within a single tampered image. To exploit such spatial relationships, we propose Proposal Contrastive Learning (PCL) for effective image manipulation detection. Our PCL consists of a two-stream architecture by extracting two types of global features from RGB and noise views respectively. To further improve the discriminative power, we exploit the relationships of local features through a proxy proposal contrastive learning task by attracting/repelling proposal-based positive/negative sample pairs. Moreover, we show that our PCL can be easily adapted to unlabeled data in practice, which can reduce manual labeling costs and promote more generalizable features. Extensive experiments among several standard datasets demonstrate that our PCL can be a general module to obtain consistent improvement. The code is available at \url{https://github.com/Sandy-Zeng/PCL}.
\end{abstract}


\section{Introduction}
\begin{figure}
  \centering
  \includegraphics[width=0.5\textwidth]{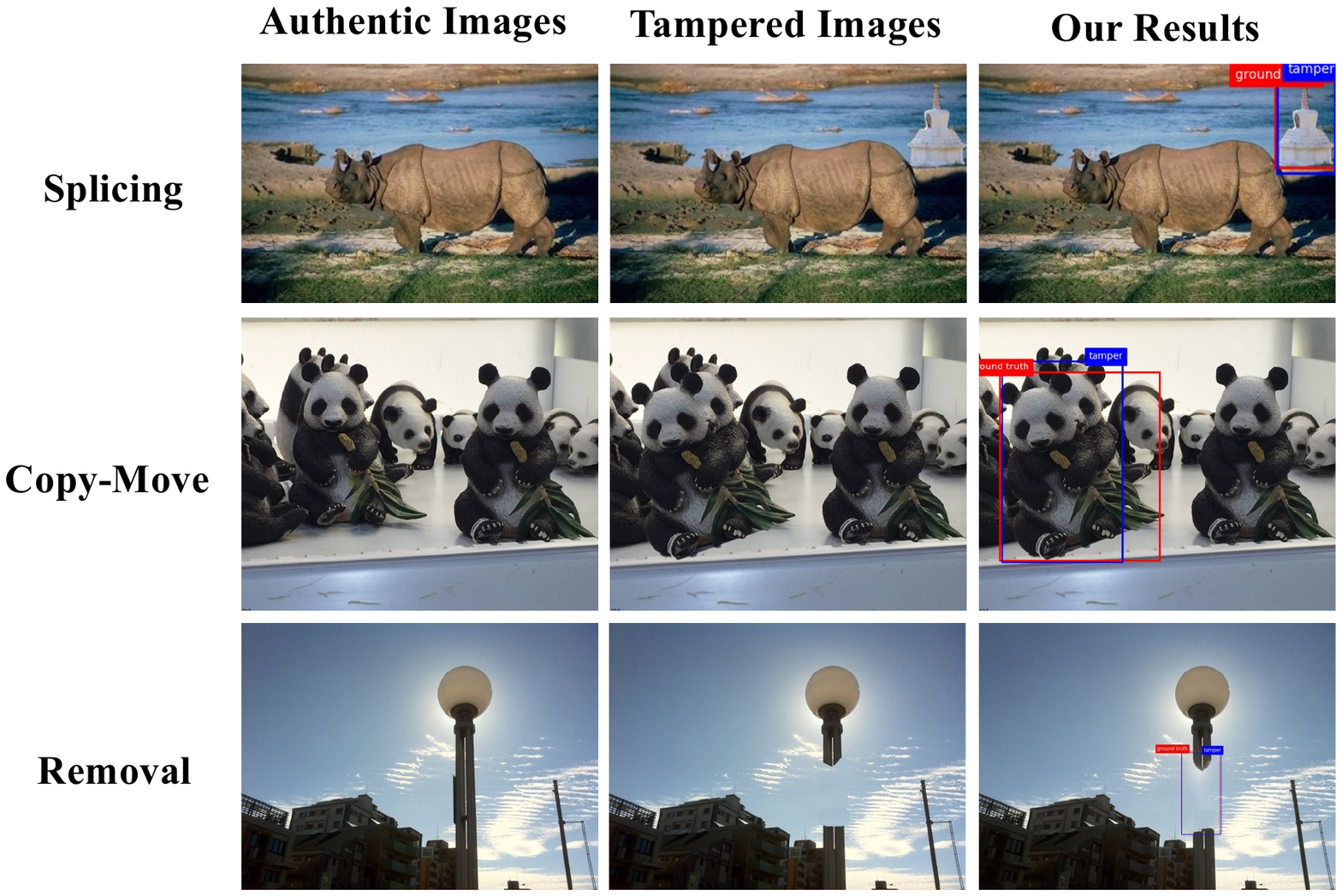}
  \caption{Tampered images with different manipulation techniques. Red boxes indicate the ground-truth bounding boxes and the blue ones are our detection results. Best viewed in color.} 
  \label{fig:introduction}
\end{figure}

\IEEEPARstart{D}{igital} images can be easily manipulated owing to easy-to-use image processing software. Malicious image manipulation, such as in fake news, insurance fraud and malicious rumors can have a great negative impact on society. 
In reality, \textit{splicing} (paste objects from one image to another), \textit{copy-move} (copy and move some elements from one region to another within one image) and \textit{removal} (erase some regions from an image followed by inpainting) are the most commonly used image manipulation techniques. 
Some examples of these image manipulations are given in Fig.~\ref{fig:introduction}. 
It is difficult to identify the tampered regions directly due to the imperceptible tampering artifacts. Faced with a considerable number of tampered images and exquisite image manipulation techniques, it is crucial to develop reliable image manipulation detection techniques.

\IEEEpubidadjcol

Early methods mainly focus on using handcrafted or predetermined features to distinguish authentic regions and tampered regions, such as doubly compressed JPEG~\cite{lin2009fast}, color filter array~\cite{ferrara2012image}, local noise features~\cite{lyu2014exposing,cozzolino2015splicebuster} and etc.
However, the handcrafted features cannot generalize well on unseen manipulation types.  
In recent years, deep learning (DL)-based methods have achieved impressive performance on image manipulation detection based on various network architectures, including convolutional neural network~\cite{chen2015median}, autoencoder~\cite{cozzolino2016single}, long short-term memory~\cite{bunk2017detection}, Faster R-CNN~\cite{zhou2018learning} and etc.
However, most existing DL-based methods mainly focus on designing sophisticated networks to extract \textit{global features} from tampered images, while they neglect the intrinsic \textit{relationships of local features} between tampered and authentic regions within a tampered image, thus hindering the discriminative ability of these networks.
For example, we commonly figure out the tampered regions by comparing the differences between normal (authentic) and abnormal (tampered) areas within an image.
We believe that such spatial correlation is helpful to improve the discriminative ability of the feature representations for manipulation detection.

\begin{figure*}[t]
\centering
\subfigure[Overall framework of our model architecture\label{fig:framework_a}]{\includegraphics[width=0.4\textwidth]{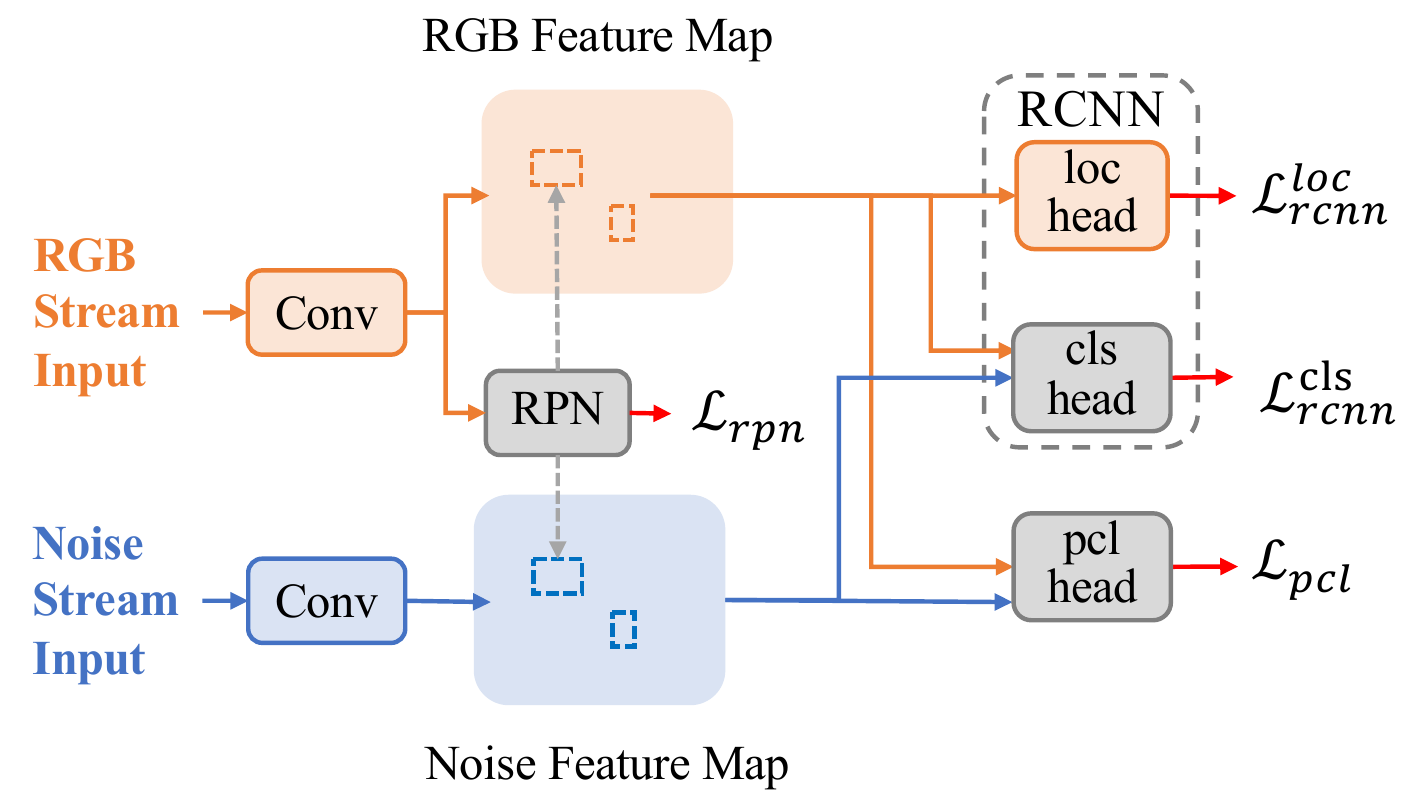}}
\hspace{5mm}
\subfigure[Illustration of the proposal contrastive learning\label{fig:framework_b}]{\includegraphics[width=0.4\textwidth]{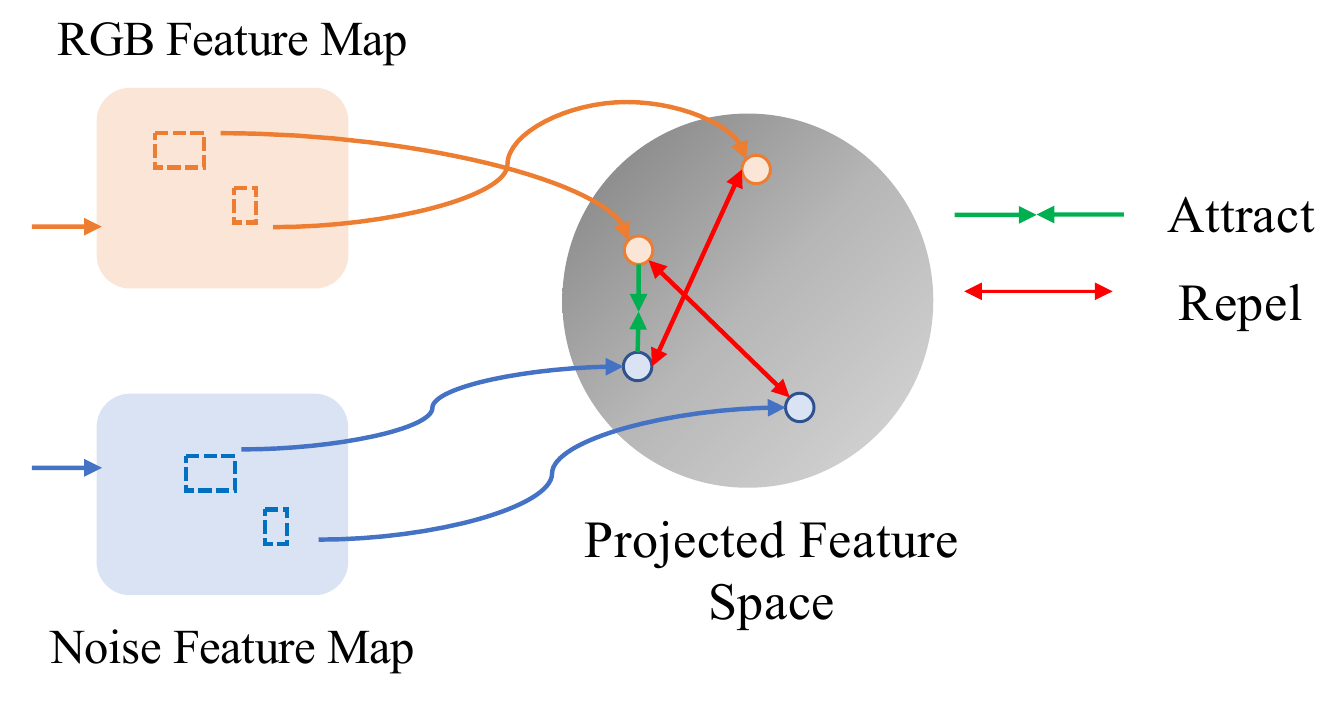}}
\caption{Our model consists of the basic detector and the Proposal Contrastive Learning (PCL) head. In the basic detector, the localization head and classification head conduct the manipulation detection task. The PCL head discriminates the proposal-based features of tampered regions and authentic regions by contrastive learning in the projected feature space.}
\label{fig:framework}
\end{figure*}

To exploit such rich spatial relationships of local features, in this paper, we propose an effective Proposal Contrastive Learning (PCL) for image manipulation detection, inspired by the success of contrastive learning in self-supervised representation learning~\cite{chen2020simple,he2020momentum}. 
As shown in Fig.~\ref{fig:framework}, our PCL module is built on a two-stream architecture to classify and locate tampered regions.
As shown by previous methods~\cite{zhou2018learning,yang2020constrained}, the noise view is beneficial to reveal the semantic-agnostic features, we adopt the two-stream architectures and construct the noise views either by SRM filters~\cite{zhou2018learning} or constrained convolutional layers~\cite{yang2020constrained}. 
\textit{Different from the conventional contrastive learning in image classification, we explore contrastive learning at proposal-level within the object detection framework.} Moreover, the two-stream features for image manipulation detection can naturally provide the positive and negative data pairs instead of constructing them with extra data augmentation as in self-supervised image classification~\cite{chen2020simple,he2020momentum}.
The relationships between tampered and authentic regions are exploited by attracting the features of positive pairs and repelling the features of negative pairs.
Moreover, our PCL can be easily applied to unlabeled data to further improve the performance, which facilitates the potential applications in practice, as numerous available tampered images without annotations are disseminated in the wild, especially on various social media platforms. The experimental results demonstrate that PCL is both effective on unlabeled data and labeled data. 
Extensive experiments among several standard datasets demonstrate that PCL can achieve consistent improvements and lead to more discriminative feature representations, showing great potential.

The main contributions of this work are summarized as follows: 
\begin{itemize}
  \item We propose a simple yet effective proposal contrastive learning module to leverage the \textit{relationships of local features} between tampered and authentic regions for discriminative representation learning. 
  \item Our method can be easily applied to unlabeled data to further improve manipulation detection performance. To the best of our knowledge, no efforts have been devoted to incorporating the massive unlabeled data for image manipulation detection.
   \item Extensive experiments on several standard datasets demonstrate the effectiveness of our proposed method on both labeled and unlabeled datasets.
\end{itemize}

\section{Related Work}

\subsection{Image Manipulation Detection}
Traditional methods for image manipulation detection mainly rely on handcrafted or predetermined features. Early work~\cite{lin2009fast} employed the statistics of DCT coefficients of doubly compressed JPEG images to distinguish the authentic and tampered regions. Color filter array was also utilized to detect the inconsistency of the tampered regions in an image~\cite{ferrara2012image}. Moreover, the local noise features introduced by the sensors and post-processing~\cite{lyu2014exposing,cozzolino2015splicebuster} and the inconsistencies of the illuminant color or lighting ~\cite{fan2015image,carvalho2015exposing} are also clues for image splicing detection. Usually, these handcrafted features are defined for a specific image manipulation type, therefore it is difficult for them to achieve good performance in practice.

Recently, extensive deep learning-based architectures are adopted to extract the tampered characteristics adaptively, leading to better generalization ability in face of different types of manipulation. Instead of relying on manual feature extraction, early proposed methods utilized convolutional neural network to determine whether an image is tampered~\cite{chen2015median}.
Besides, long short-term memory (LSTM) based network was also employed to classify and locate the tampered regions~\cite{bunk2017detection}. The two-stream network utilized an RGB stream and a noise stream to learn rich features for image manipulation detection~\cite{zhou2018learning}, which relies on the modern object detection models. In other work, deep semantic segmentation models, like fully convolutional networks (FCN)~\cite{salloum2018image}, hybrid models~\cite{bappy2019hybrid,wu2019mantra} were also employed for image manipulation detection.
In CR-CNN~\cite{yang2020constrained}, they proposed a coarse-to-fine architecture based on Mask R-CNN, which also incorporates the constrained convolutional layer and attention mechanism. SNIS~\cite{chen2022snis} propose a signal noise separation module to separate tampered region for post-processing forgery detection. Instead of designing more sophisticated network structures, in this work, we focus on (i) how to obtain more discriminative feature representations, (ii) how to make full use of unlabeled data to alleviate the data-scarce problem.

Moreover, there are other studies utilizing the patch consistency within the images to identify the tampered regions, such as~\cite{huh2018fighting,cozzolino2019noiseprint,mayer2019forensic}. They rely on different camera attributes to identify tampered regions, such as EXIF~\cite{huh2018fighting}, Noiseprint~\cite{cozzolino2019noiseprint} and other adaptive features learned from data directly~\cite{mayer2019forensic}. Thus these methods require data with meta-information about camera, which is often not available. While we use the manipulated images with only pixel information. Previous work explicitly learns the patch consistency by dividing images into multiple overlapped patches and performing patch pairings to identify differences. While our method implicitly models patch consistency through contrastive learning at proposal level. Therefore, our method is different from previous work in terms of the data condition and the utilization of patch consistency.

\subsection{Contrastive Learning}
Recently, contrastive learning has made important progress in self-supervised learning~\cite{he2020momentum,chen2020simple,tian2019contrastive}, which aims to learn effective representations without human supervision, i.e., from unlabeled data. 
In addition, the contrastive representation has been shown to play a beneficial role in other tasks, like semi-supervised learning~\cite{zhai2019s4l}, supervised image recognition~\cite{khosla2020supervised} and unsupervised video representation learning~\cite{chen2022consistent}. Instead of defining loss functions to measure the difference between predictions and fixed targets, contrastive losses measure the similarities of data pairs in representation space. The contrastive losses push apart dissimilar sample pairs and pull together similar sample pairs. How choose positive and negative data pairs is extremely important in contrastive learning. In self-supervised image recognition, the most common practice is to treat different data augmentations of the same sample as positive data pairs, and the augmentations from other images are regarded as negative data pairs. 
Others like different modalities~\cite{chung2019perfect,morgado2020audio}, different feature representations~\cite{tian2019contrastive} can also provide positive data pairs.

In this work, we explore contrastive learning at proposal-level for image manipulation detection. Relying on two-stream inputs (RGB and noise), the positive and negative sample pairs are constructed. Our proposal contrastive learning is employed to enhance the feature representations under both the supervised paradigm and the semi-supervised paradigm.
{\color{black} The concurrent work~\cite{yin2022contrastive,pan2022auto} also utilizes the tool ``contrastive learning" to enhance image manipulation detection. However, there are many differences between our work with them: 
(i) The concurrent work is based on the semantic segmentation framework, which must require per-pixel annotations. While our model is based on the object detection framework and can be trained with the bounding-box annotations, which are much easier and cheaper than the per-pixel annotations (35$\times$ -- 85$\times$ improvement in annotation speed~\cite{kulharia2020box2seg}).
(ii) Also for the above reason, our work mainly outputs the predicted bboxes and has marginal advantages in terms of pixel-level evaluation metrics. However, in many real applications, there is actually no need to discover every tampered pixel. Instead, the complete and accurate bounding boxes are enough. Thus, the detection-framework-based methods are also valuable, which can meet the need and do not require costly annotations.
(iii) Our contrastive learning module allows the unlabeled data to participate in training and brings additional gains, which is important and urgently required in the real world. However, the concurrent work does not offer this capability (their contrastive learning mechanism must rely on labeled information).
}

\section{Methodology}

\subsection{Basic Detector}
\label{sec: framework}

We first introduce the widely used basic detector~\cite{zhou2018learning} for manipulation detection.

\textbf{Overall Framework.} As shown in Fig.~\ref{fig:framework_a}, the basic model consists of two streams: an RGB stream and a noise stream. The RGB stream operates on the original RGB image, and the noise stream takes the noise feature map processed by SRM filters~\cite{fridrich2012rich} or constrained convolutional layers~\cite{yang2020constrained} as input. As indicated by~\cite{zhou2018learning}, the noise stream can pay more attention to the tampered characteristics rather than the semantic image content. Faster R-CNN is adopted to the design of the two-stream network architecture. The RGB stream and noise stream have independent backbone networks (e.g., ResNet-101). These two streams share one Region Proposal Network (RPN), which is elaborated below. Then, the proposal features extracted by the two streams are fed into the Region-based CNN (RCNN), which contains a classification head and a localization head, see details below.

\textbf{Region Proposal Network (RPN).} The RPN only relies on the RGB stream to learn and generate region proposals. The noise stream directly borrows the RPN module from the RGB stream. Formally, the training loss for the RPN consists of a classification loss $\mathcal{L}_{rpn}^{cls}$ (e.g., the cross-entropy loss) and a localization loss $\mathcal{L}_{rpn}^{loc}$ (e.g., the smooth-$L_1$ loss) as follows,
\begin{equation}
\label{equ: rpn}
\begin{aligned}
\mathcal{L}_{rpn} =  \mathcal{L}_{rpn}^{cls}(\{\mathbf{p}_k\}, \{\mathbf{p}^*_{k}\})
+ \lambda_1 \mathcal{L}_{rpn}^{loc}(\{\mathbf{l}_k\}, \{\mathbf{l}^*_{k}\}), 
\end{aligned}
\end{equation}
where $\mathbf{p}_k$ and $\mathbf{p}^*_{k}$ denote the probability of anchor $k$ being a potential tampered region, the corresponding true label. $\mathbf{l}_k$ and $\mathbf{l}^*_{k}$ are the predicted 4-dimensional bounding box coordinates for anchor $k$ and the corresponding ground-truth coordinates respectively. $\lambda_1$ is the hyper-parameter to balance the two losses.

\textbf{Region-based CNN (RCNN).} After RPN, features are cropped from the RGB and noise feature maps for each proposal. Next, the features are resized to a fixed size through an RoI Pooling layer and fed into RCNN, which is responsible for giving the class probabilities for each proposal and refining the corresponding boxes. The RCNN is composed of a classification head and a localization head. In the classification head, the RoI features $\mathbf{h}^{r}$ and $\mathbf{h}^{n}$ generated from RGB stream and noise stream are combined via bilinear pooling ($BiPool$), which can preserve the spatial information and fuse the features from two streams naturally. The fused features for classification are given by
\begin{equation}
  \mathbf{h}^{f} = BiPool \ (\mathbf{h}^{r},\ \mathbf{h}^{n}).
\end{equation}
Considering the RGB information is more effective for localization, only the features of RGB stream are provided to the localization head, as shown in Fig.~\ref{fig:framework_a}. Overall, the training loss for RCNN is formulated as
\begin{equation}
\label{equ: bb}
\begin{aligned}
\mathcal{L}_{rcnn} =  \mathcal{L}_{rcnn}^{cls}(\{\mathbf{p}_i\}, \{\mathbf{p}^*_{i}\})
+ \lambda_2 \mathcal{L}_{rcnn}^{loc}(\{\mathbf{l}_i\}, \{\mathbf{l}^*_{i}\}), 
\end{aligned}
\end{equation}
where $\mathcal{L}_{rcnn}^{cls}$ and $\mathcal{L}_{rcnn}^{loc}$ are the classification and localization loss, $\mathbf{p}_i$ denotes the probability of proposal $i$ being a potential tampered region and $\mathbf{p}^*_{i}$ is the ground truth label. $\mathbf{l}_i$ and $\mathbf{l}^*_{i}$ are the predicted 4-dimensional bounding box coordinates for proposal $i$ and the corresponding ground-truth coordinates respectively. $\lambda_2$ is the hyper-parameter to balance the two losses. $\mathcal{L}_{rcnn}^{cls}$ is related to the fused features $\mathbf{h}^{f}$, and $\mathcal{L}_{rcnn}^{loc}$ is associated to the RGB features $\mathbf{h}^{r}$ alone.

\subsection{Proposal Contrastive Learning}
\label{sec: pcl}

A powerful representation is the one that models view-invariant factors~\cite{tian2019contrastive}, inspired by that, we would like to improve the consistency of features from the two streams towards manipulation characteristics, and enhance the distinction of features between authentic regions and tampered regions. 

Naturally, we propose a Proposal Contrastive Learning (PCL) module that draws the features between positive pairs closer and pulls away the negative pairs. Specifically, as shown in Fig.~\ref{fig:framework_a} and Fig.~\ref{fig:framework_b}, we add a PCL head with proposal contrastive learning loss $\mathcal{L}_{pcl}$ to the original basic detector, which can interact between the features extracted from the two streams and lead to more discriminative features. In the PCL head, we adopt two fully connected layers as projection heads to map the RGB features $\mathbf{h}^{r}$ and noise features $\mathbf{h}^{n}$ to a low-dimensional feature space, which facilitates contrastive learning~\cite{chen2020simple}. Formally, given the projection heads $g^{r}(\cdot)$ and $g^{n}(\cdot)$ for the two streams, we project the RoI features to the low dimensional vectors by 
\begin{equation}
\mathbf{z}_i^r = g^{r}(\mathbf{h}_{i}^{r}) = W_2^{r} \sigma(BN(W_1^{r} \mathbf{h}_{i}^{r})),
\end{equation}
\begin{equation}
\mathbf{z}_i^n = g^{n}(\mathbf{h}_{i}^{n}) = W_2^{n} \sigma(BN(W_1^{n} \mathbf{h}_{i}^{n})),
\end{equation}
where $BN$ represents batch normalization, $W_1$ and $W_2$ are the parameters in projection head, $\sigma$ is the ReLU non-linearity, $\mathbf{z}_i^r$ and $\mathbf{z}_i^n$ are the projected feature representations from the RGB stream and the noise stream for proposal $i$. 

Then, we would like to construct positive and negative pairs among the projected feature representations $\{\mathbf{z}_i^r\}$ and $\{\mathbf{z}_i^n\}$, which is crucial for contrastive learning. As illustrated in Fig.~\ref{fig:construction of pairs}, the vanilla strategy views the two representations $\{(\mathbf{z}_i^r, \mathbf{z}_i^n)\}$ of proposal $i$ as positive pairs, and views others pairs $\{(\mathbf{z}_i^r, \mathbf{z}_j^n), j \neq i\}$ as negative pairs, which is named fully contrastive loss, dubbed as PCL-FCL. However, recent work shows that the false negative pairs prevent contrastive learning from better representations~\cite{huynh2020boosting}. Therefore, it is essential to mitigate the negative effects of false negative pairs. We achieve this by establishing positive and negative pairs thoughtfully. We describe the construction of positive/negative pairs for labeled data and unlabeled data in Sec.~\ref{sec: super pcl} and Sec.~\ref{sec: semi-super pcl} respectively.

\subsection{Proposal Contrastive Learning on Labeled Data}
\label{sec: super pcl}
\begin{figure}[t]
\centering
\includegraphics[width=0.5\textwidth]{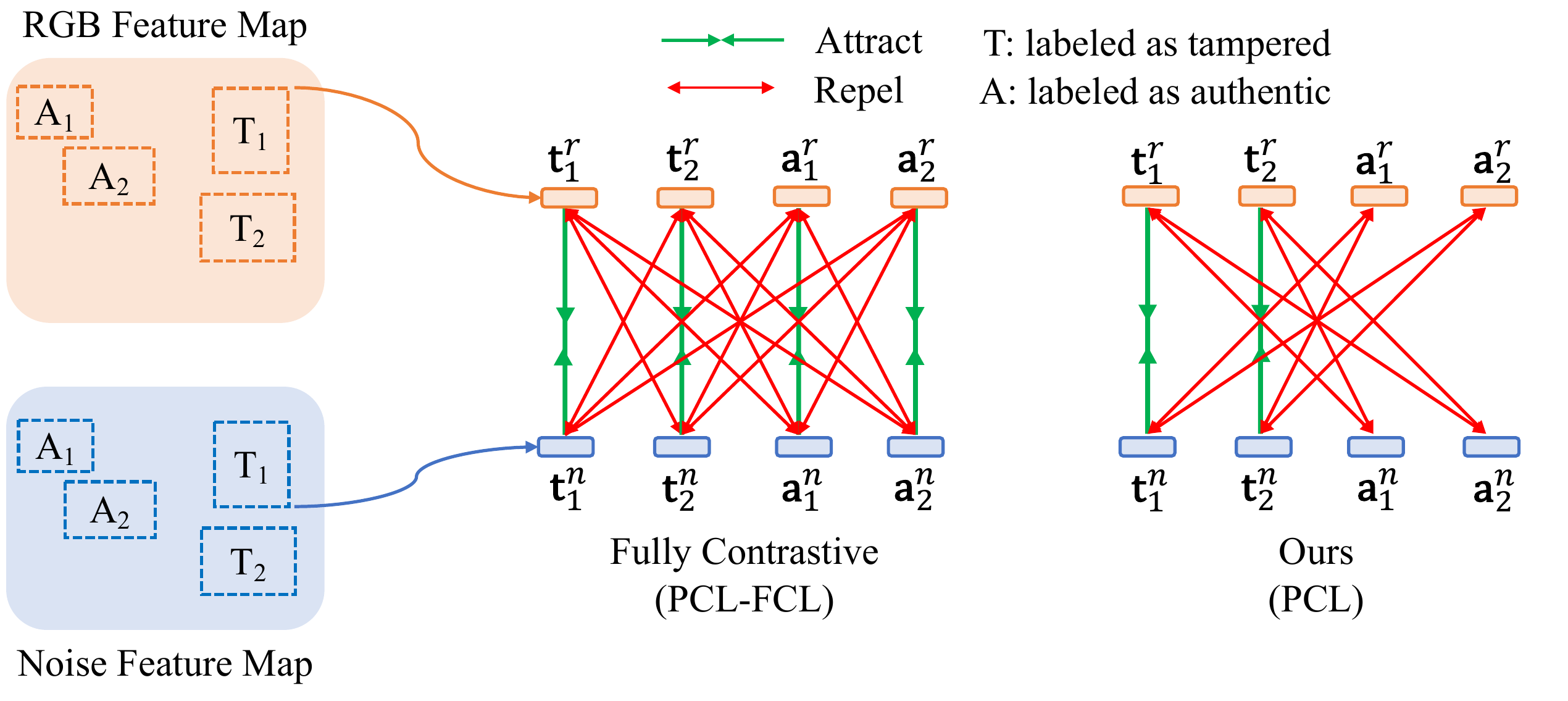}
\caption{Construction of positive/negative pairs. Positive pairs between RGB map and noise map are attracted while negative pairs are repelled. (Best viewed in color.)}
\label{fig:construction of pairs}
\end{figure}

When applying PCL loss on labeled data, we construct the positive and negative pairs based on the real annotations. Specifically, if the Intersection over Union (IoU) between the proposal and ground-truth bounding box exceeds the threshold $\epsilon$, it is labeled as ``tampered", otherwise marked as ``authentic". The projected feature representations $\{ \mathbf{z}^r_i \}$, $\{ \mathbf{z}^n_i \}$ from RGB stream and noise stream for the \underline{t}ampered proposals are rewritten as $\{\mathbf{t}_i^{r}\}$ and $\{\mathbf{t}_i^{n}\}$ respectively, for the \underline{a}uthentic proposals are rewritten as $\{\mathbf{a}_i^{r}\}$ and $\{\mathbf{a}_i^{n}\}$. Correspondingly, we construct the positive pairs $\{(\mathbf{t}_i^{r}, \mathbf{t}_i^{n})\}$, the negative pairs $\{(\mathbf{t}_i^{r}, \mathbf{a}_j^{n})\}$ and $\{(\mathbf{t}_i^{n}, \mathbf{a}_j^{r})\}$, where the RGB features and noise features for the same tampered proposals are constructed as positive pairs, the features of tampered proposals and authentic proposals are assigned as negative pairs. 
The illustration is shown in Fig.~\ref{fig:construction of pairs}.
Then, following the InfoNCE loss \cite{oord2018representation}, the proposed PCL loss is defined as
\begin{equation}
\label{equ: pcl}
\begin{aligned}
    &\mathcal{L}_{pcl} = \\ 
    &- \frac{1}{N}
    \sum_{i} \log 
    \frac{
    \exp(s(\mathbf{t}_i^{r}, \mathbf{t}_i^{n}, \tau))
    }
    {
    \exp(s(\mathbf{t}_i^{r}, \mathbf{t}_i^{n}, \tau)) +  
    \sum_{j} \exp(s(\mathbf{t}_i^{r}, \mathbf{a}_j^{n}, \tau))
    } \\
    & - \frac{1}{N}
    \sum_{i} \log 
    \frac{
    \exp(s(\mathbf{t}_i^n, \mathbf{t}_i^{r}, \tau))
    }
    {
    \exp(s(\mathbf{t}_i^{n}, \mathbf{t}_i^{r}, \tau)) + 
    \sum_{j} \exp(s(\mathbf{t}_i^{n}, \mathbf{a}_j^{r}, \tau))
    }, 
\end{aligned}
\end{equation}
where $s(\mathbf{u}, \mathbf{v}, \tau) = \frac{\mathbf{u}^T \mathbf{v}}{\tau \| \mathbf{u} \| \| \mathbf{v} \|}$ stands for the similarity between two inputs $\mathbf{u}$ and $\mathbf{v}$, $\tau$ is the hyper-parameter of temperature, $N$ is the number of positive pairs. Here, the similarity of features is measured by dot product. The PCL loss maximizes the agreement between the RGB and noise features of tampered regions, which helps reveal the view-invariant factors, tampered characteristics here; on the other hand, it can repel the features of authentic regions. 

Equ.~\eqref{equ: pcl} is composed of two terms. The first term (denoted as $\mathcal{L}_{pcl-rgb}$) treats the RGB features from tampered regions $\{\mathbf{t}_i^{r}\}$ and the noise features from authentic regions $\{\mathbf{a}_j^{n}\}$ as negative pairs, dubbed as PCL-RGB. The second term (denoted as $\mathcal{L}_{pcl-noise}$) regards the noise features from tampered regions $\{\mathbf{t}_i^{n}\}$ and the RGB features from authentic regions $\{\mathbf{a}_j^{r}\}$ as negative pairs oppositely, dubbed as PCL-Noise. In Sec. \ref{sec: ablation study}, we conduct an ablation study on different loss functions and empirically demonstrate that we can obtain better performance when using the PCL loss. Overall, the total loss function for image manipulation detection with labeled data is given by
\begin{equation}
\label{equ: total}
    \mathcal{L}^l = \mathcal{L}_{rpn} + \mathcal{L}_{rcnn} + \beta \mathcal{L}_{pcl}, 
\end{equation}
where $\beta$ balances the proxy contrastive learning task and the original tasks.

\subsection{Proposal Contrastive Learning on Unlabeled Data}
\label{sec: semi-super pcl}
We have introduced PCL in supervised learning regime, where the construction of positive/negative pairs is based on the ground-truth bounding boxes. However, labeling tampered images is time-consuming and labor-intensive. Thus, in addition to labeled data, we perform PCL on unlabeled data, which can incorporate a lot of ``free" data into training, leading to enriched feature representations.

\begin{figure}[t]
\centering
\subfigure[\label{fig1:a}]{\includegraphics[width=0.45\columnwidth]{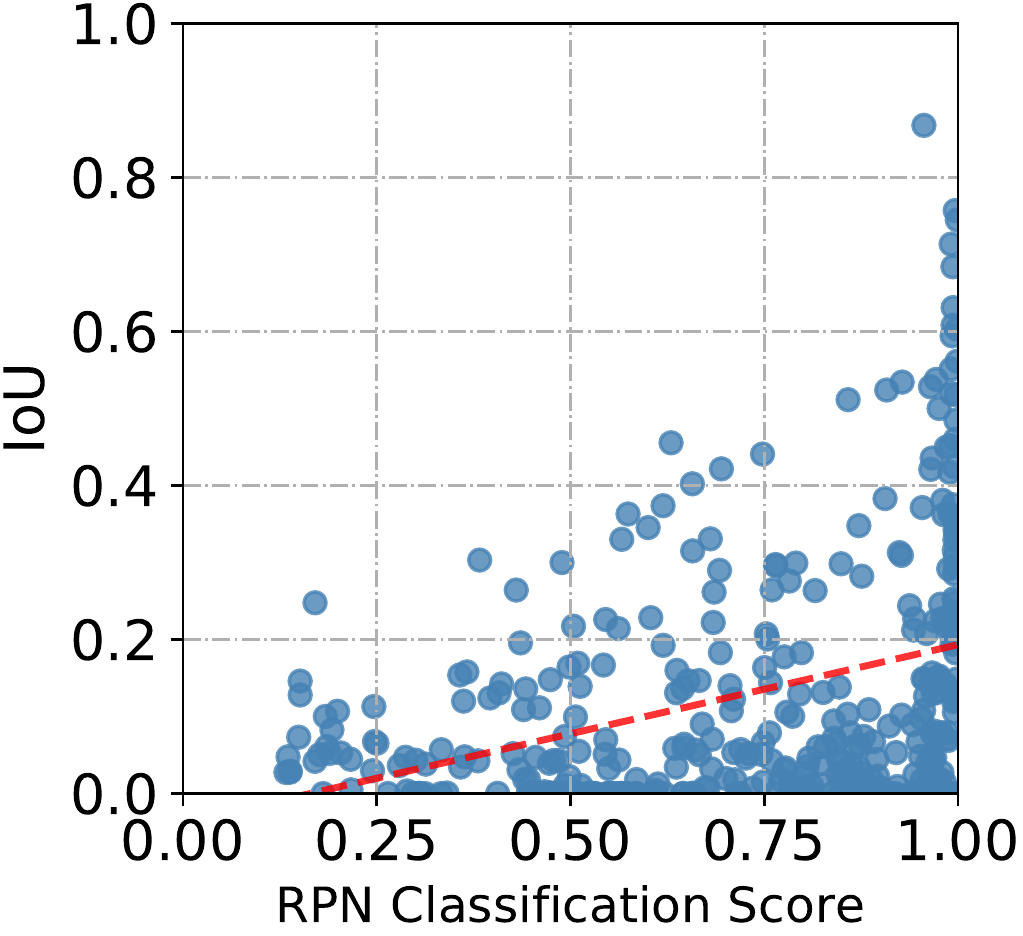}} \hspace{2mm}
\subfigure[\label{fig1:b}]{\includegraphics[width=0.45\columnwidth]{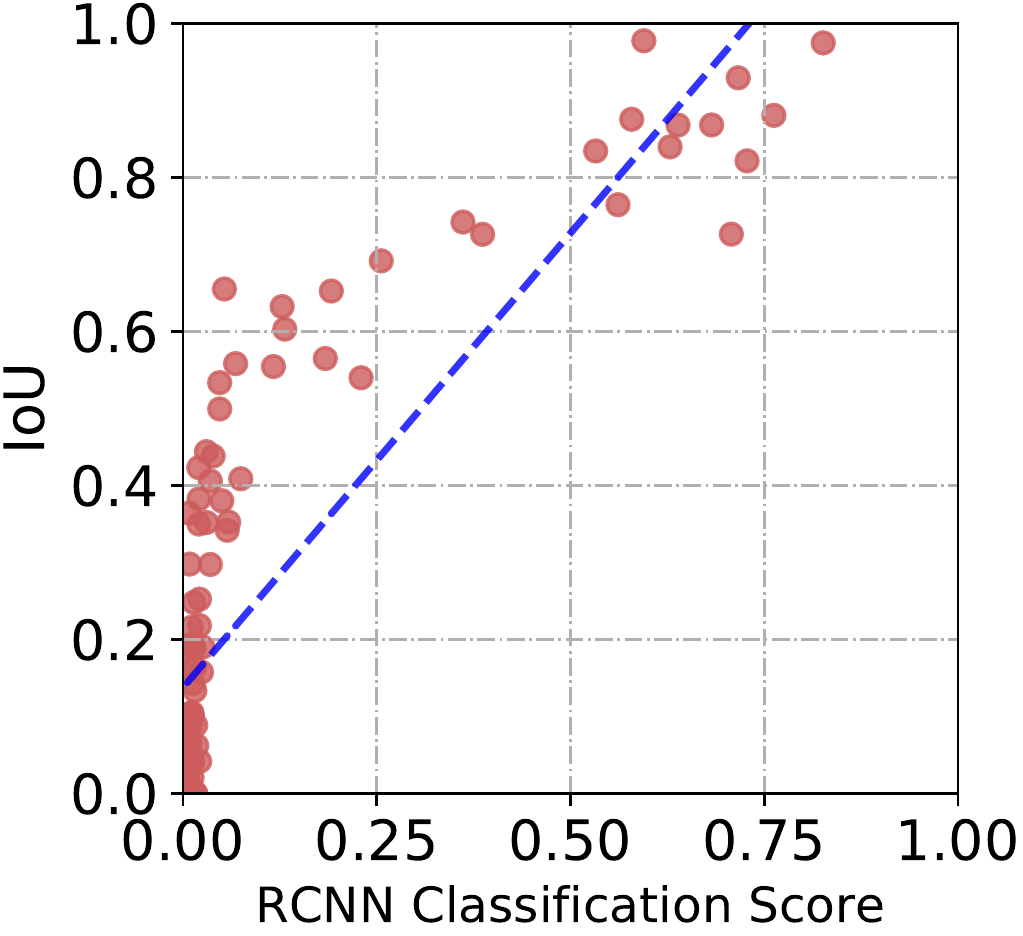}}
\caption{The correlation of IoU (between proposals and ground-truth boxes) and RPN scores (left), RCNN scores (right). The dashed line in each figure is the linear fitting curve. The PCC of IoU and RPN scores is 0.57 while for IoU and RCNN scores is 0.83.}
\label{fig:iou_scores}
\end{figure}

For unlabeled data, the ground-truth boxes are not available to identify the positive and negative pairs like for labeled data, thus we turn to rely on the prediction results to construct sample pairs. There are two prediction scores related to a proposal: the score in RPN and that in RCNN. Here, we examine two different sampling strategies: (i) one depends on the predicted classification scores of RPN; (ii) the other depends on the predicted classification scores of RCNN. We hope that the prediction scores can reflect whether the proposals are related to tampered regions, which can help to select more suitable sample pairs. In Fig.~\ref{fig:iou_scores}, we draw the relationship of IoU (between the proposals and ground-truth boxes) and the prediction scores in RPN (Fig.~\ref{fig1:a}) and RCNN (Fig.~\ref{fig1:b}) respectively. It can be observed in Fig.~\ref{fig1:a} that the proposals distribute uniformly with RPN scores due to the RPN usually guarantees recall rather than precision. While RCNN scores are more correlated with IoU as depicted in Fig.~\ref{fig1:b}. We also calculate the Pearson Correlation Coefficient (PCC)~\cite{goh2007human} to indicate the linear correlation of IoU and prediction scores. The PCC of IoU and RPN scores is 0.57, while for IoU and RCNN scores is 0.83, which indicates that RCNN scores are more reliable. In Sec.~\ref{sec: ablation study}, we also conduct an ablation study to show the detection performance with RPN and RCNN scores as guidance for unlabeled data.

Based on the above observations, we apply the RCNN scores as guidance to select positive and negative pairs. The proposal $i$ is considered as ``tampered" when the RCNN score is larger than the threshold $\delta$, otherwise labeled as ``authentic". Now, like in Sec.~\ref{sec: super pcl}, we can divide the projected feature representations $\{ \mathbf{z}^r_i \}$, $\{ \mathbf{z}^n_i \}$ of unlabeled data into $\{\mathbf{t}_i^r\}$, $\{\mathbf{t}_i^n\}$ and $\{\mathbf{a}_i^r\}$, $\{\mathbf{a}_i^n\}$. Then, the same strategy can be applied to construct positive and negative pairs. Finally, we can compute the contrastive loss $\mathcal{L}^u_{pcl}$ for unlabeled data by Equ.~\eqref{equ: pcl}. 

{\color{black} Finally, the labeled data and unlabeled data are utilized together and contributed to different loss functions, \textit{i.e.}, $\mathcal{L}^l$ and $\mathcal{L}^u_{pcl}$ respectively.}
Overall, the total loss function is given by
\begin{equation}
\label{equ: semi loss}
    \mathcal{L}^{lu} = \mathcal{L}^l + \omega(m) \cdot \mathcal{L}^u_{pcl},
\end{equation}
where $\mathcal{L}^l$ is computed as Equ.~\eqref{equ: total} on labeled data, $\mathcal{L}^u_{pcl}$ is computed on unlabeled data weighted by a weight function $\omega(m)$ which increases along as the training schedule. Following~\cite{laine2016temporal}, the weight function is defined as $\omega(m)=\exp(-5\cdot \max(1-\frac{m}{M}, 0)^2)$, where $m$ is the number of current iteration, and $M$ is the number of warm up iterations. Since the prediction results of RCNN are not reliable at the beginning of network training, a smaller loss weight is adopted for unlabeled data at first.

\section{Experiments}
We conduct experiments on several standard image manipulation datasets and compare our method with other counterparts to demonstrate the effectiveness of proposal contrastive learning. Moreover, as our method can be applied to unlabeled data, we further incorporate unlabeled data into training to improve the performance.

\subsection{Experimental Settings.} 
\subsubsection{Implementation Details}
The proposed model is trained in an end-to-end manner. During training, image flipping and gaussian noise are used for data augmentation. The input images are resized such that their shorter side equals 600 pixels. For anchors, we use four scales with box areas of $8^2,16^2,32^2$and $64^2$ pixels, and three aspect ratios of 1:1, 1:2, and 2:1. The anchor scales are smaller than that in original Faster R-CNN~\cite{ren2015faster} because we focus on regions that are likely to be manipulated instead of meaningful objects in the original Faster R-CNN. The batch size of RPN proposals is set to 32 for training and set to 100 for testing. {\color{black} Following the default and common setting~\cite{ren2015faster}}, we set IoU threshold to 0.7 for potential positive regions and 0.3 for negative regions for anchors in RPN network. We use the minimum boxes containing ground-truth masks as ground-truth bounding boxes for our model training. 
The hyper-parameter $\lambda_1$ in Equ.~\eqref{equ: rpn} is set to 10, and $\lambda_2$ in Equ.~\eqref{equ: bb} is set to 1. We set IoU threshold $\epsilon$ for labeled data and predicted scores threshold $\delta$ for unlabeled data to 0.5 to distinguish tampered regions from authentic regions. Temperature $\tau$ and $\beta$ in Equ.~\eqref{equ: pcl} are set to 0.1 and 0.05 in default setting. We pre-train our model with COCO synthetic dataset~\cite{zhou2018learning} for 110K steps. The warm up iterations $M$ for unlabeled data is set to 60K steps. The learning rate is initially set to $1e-3$ and reduced to $1e-4$ after 40K steps. 
{\color{black} All the experiments are conducted in a single GTX 1080Ti.
}

\subsubsection{Evaluation Metrics}
We use AP$_{50}$ and AP$_{50-95}$, the common metrics~\cite{lin2014microsoft,everingham2010pascal}, to compare our PCL with other detection-based methods.
{\color{black} AP$_{50}$ corresponds to the average precision at IoU threshold of 0.5, and AP$_{50-95}$ means the average AP at IoU thresholds from 0.5 to 0.95 with a step size of 0.05.}
{\color{black} To evaluate the image-level manipulation detection (recognition) performance, we use image-level Area Under Curve (AUC) score and F1 score}. For the pixel-level localization performance, we use pixel-level F1 score and AUC for comparison with those segmentation-framework-based methods following previous work~\cite{salloum2018image}.

\subsubsection{Evaluation Datasets}
We use the same COCO synthetic dataset in~\cite{zhou2018learning} to pre-train and evaluate our models. The proposed methods are compared with others on four public image manipulation benchmarks:  NIST16~\cite{guan2019mfc}, CASIA~\cite{dong2010casia,dong2013casia}, COVERAGE~\cite{wen2016coverage},  Columbia~\cite{ng2009columbia} and IMD20~\cite{novozamsky2020imd2020}. The training and testing split protocol of each benchmark follows~\cite{zhou2018learning}. The details of the evaluation datasets are presented as below:
\begin{itemize}
    \item \textbf{COCO synthetic dataset} is created from COCO \cite{lin2014microsoft} by randomly selecting objects from one image with the segmentation mask and splicing to other images which is used for model pre-training. The dataset contains 11K manipulated images with the ground-truth tampered masks.
    \item \textbf{NIST16} is a challenging dataset of Nimble Challenge Evaluation 2016. It contains three tampering types. Masks are provided to indicate where manipulation takes place. The dataset is split into two groups: 404 images for fine-tuning and 160 images for evaluation.
    \item \textbf{CASIA} contains splice and copy-move tampered images. By carefully selecting the tampered areas and using post-processing methods like blurring, the tampered images are as much as realistic to human eyes. We obtain the ground-truth masks by subtracting tampered images from their original images. We use CASIA 2.0 with 5,123 images for fine-tuning and CASIA 1.0 with 921 images for evaluation.
    \item \textbf{COVERAGE} is a small copy-move dataset. A tampered image is created by copying a small object from a duplicated region and pasting it onto the same image. Ground-truth masks are provided. We use 75 images for fine-tuning and 25 images for evaluation.
    \item \textbf{Columbia} is a small dataset containing 180 splicing images. Ground-truth marks are provided. We directly use the pre-trained model to evaluate Columbia without fine-tuning as the same in~\cite{zhou2018learning}.
    \item \textbf{IMD20} consists of 2,010 real-life manipulated images collected from Internet, it contains all three types of manipulation.
\end{itemize}

\subsection{Ablation Study}
\label{sec: ablation study}
In this section, we analyze different components of our proposed method including loss functions, different noise views, effects of hyper-parameter $\beta$ and proposal sampling strategies for unlabeled data. Experiments in this section are conducted on COCO synthetic dataset unless otherwise specified. 

\subsubsection{Different Loss Functions}
Firstly, we test different loss functions e.g. PCL-RGB, PCL-Noise, PCL-FCL and PCL, which reveals different strategies for constructing positive and negative pairs as discussed in Sec.~\ref{sec: super pcl}. PCL-RGB and PCL-Noise indicate that we only apply the $\mathcal{L}_{pcl-rgb}$ or $\mathcal{L}_{pcl-noise}$  in Equ.~\eqref{equ: pcl} respectively. PCL-FCL means the utilization of vanilla fully contrastive strategy to construct the positive and negative pairs as illustrated in Fig.~\ref{fig:construction of pairs}. The baseline method is RGB-N \cite{zhou2018learning} which utilizes SRM filters to construct the noise view. From Table~\ref{tab:pair construction},  we can observe that model trained with PCL can improve AP$_{50}$ of baseline from 74.1\% to 78.0\% and enhance AP$_{50-95}$ from 50.3\% to 53.2\%, which indicates that contrastive learning leads to better feature representations for image manipulation detection. The comparison of PCL-RGB, PCL-Noise and PCL indicates that bilateral contrastive learning is necessary for better performance. In the case of PCL-FCL, AP$_{50}$ is 76.3\%, outperforming the baseline by 2.1\%. However, it is still inferior to PCL (76.3\% v.s. 78.0\%), which demonstrates that selecting suitable positive and negative sample pairs plays an important role in contrastive learning.

\begin{table}[t]
\caption{Ablation study on different loss functions. PCL-RGB, PCL-Noise, PCL-FCL, PCL are three loss function with different pair construction strategies. 
AP$_{50}$ (\%) and AP$_{50-95}$(\%) are reported.}
\label{tab:pair construction}
\centering
\begin{tabular}{l|cc}
\toprule
Loss Function & AP$_{50}$ & AP$_{50-95}$\\
\midrule
Baseline & 74.1 & 50.3 \\
\midrule
PCL-RGB & 76.2 & 51.5 \\
PCL-Noise & 71.6 & 48.9 \\
PCL-FCL  & 76.3 & 52.6 \\
\midrule
PCL(Ours) & \textbf{78.0} & \textbf{53.2}\\
\bottomrule
\end{tabular}
\end{table}

\subsubsection{Different Noise Views}
Most image manipulation detection methods aim to exploit the different noise distribution between authentic and tampered parts to detect tampered images~\cite{zhou2018learning,yang2020constrained}. 
The noise views of input images are generated by the pre-defined filters such as SRM filters in RGB-N~\cite{zhou2018learning} or the trainable counterparts such as the constrained convolutional layers denoted as Constrained Conv in CR-CNN~\cite{yang2020constrained}.
In our method, we adopt the two-stream architecture, for the noise stream, we evaluate the two choices of noise views in this section. The experimental results are summarized in Table~\ref{tab: noise view}. We adopt RGB-N to indicate using SRM filters for noise view construction, and RGB-C to represent the one with Constrained Conv as the input layer. The experimental results indicate that the Constrained Conv is better than SRM filters for noise feature extraction. For both two choices of noise views, when combined with PCL, the performance can be improved by 1\%-3\%, which shows that PCL can adaptively extract the valuable features from the two different views. 
Moreover, we compare RGB-C (a two-stream architecture) with CR-CNN (a single-stream architecture with single noise feature extraction). The CR-CNN results are reproduced with their open-source code. It can be observed that RGB-C outperforms CR-CNN in AP$_{50-95}$ on COCO synthetic dataset (53.4\% v.s. 52.9\%). We can continuously improve the performance by combining RGB-C with PCL, e.g. AP$_{50}$ is 79.4\% and AP$_{50-95}$ is 55.1\%, which shows that two-stream architectures can provide more diverse feature representations than the single-stream ones. 

\begin{table}[t]
\caption{Experiments on different noise views with PCL on COCO synthetic dataset. AP$_{50}$ (\%) and AP$_{50-95}$(\%) are reported.}
\label{tab: noise view}
\centering
\begin{tabular}{lc|cc}
\toprule
Methods & Noise View & AP$_{50}$ & AP$_{50-95}$\\
\midrule
RGB-N~\cite{zhou2018learning} & \multirow{2}*{SRM filter} & 74.1 & 50.3 \\
+PCL(Ours) &  & \textbf{78.0} & \textbf{53.2} \\
\midrule
CR-CNN~\cite{yang2020constrained} & Constrained Conv & 78.3 & 52.9 \\
\midrule 
RGB-C & \multirow{2}*{Constrained Conv} & 78.1 & 53.4 \\ 
+PCL(Ours) & & \textbf{79.4} & \textbf{55.1} \\
\bottomrule
\end{tabular}
\end{table}

\subsubsection{Effect of $\beta$}
\label{sec: effect of beta}
\begin{figure}[t]
\centering
\includegraphics[width=0.3\textwidth]{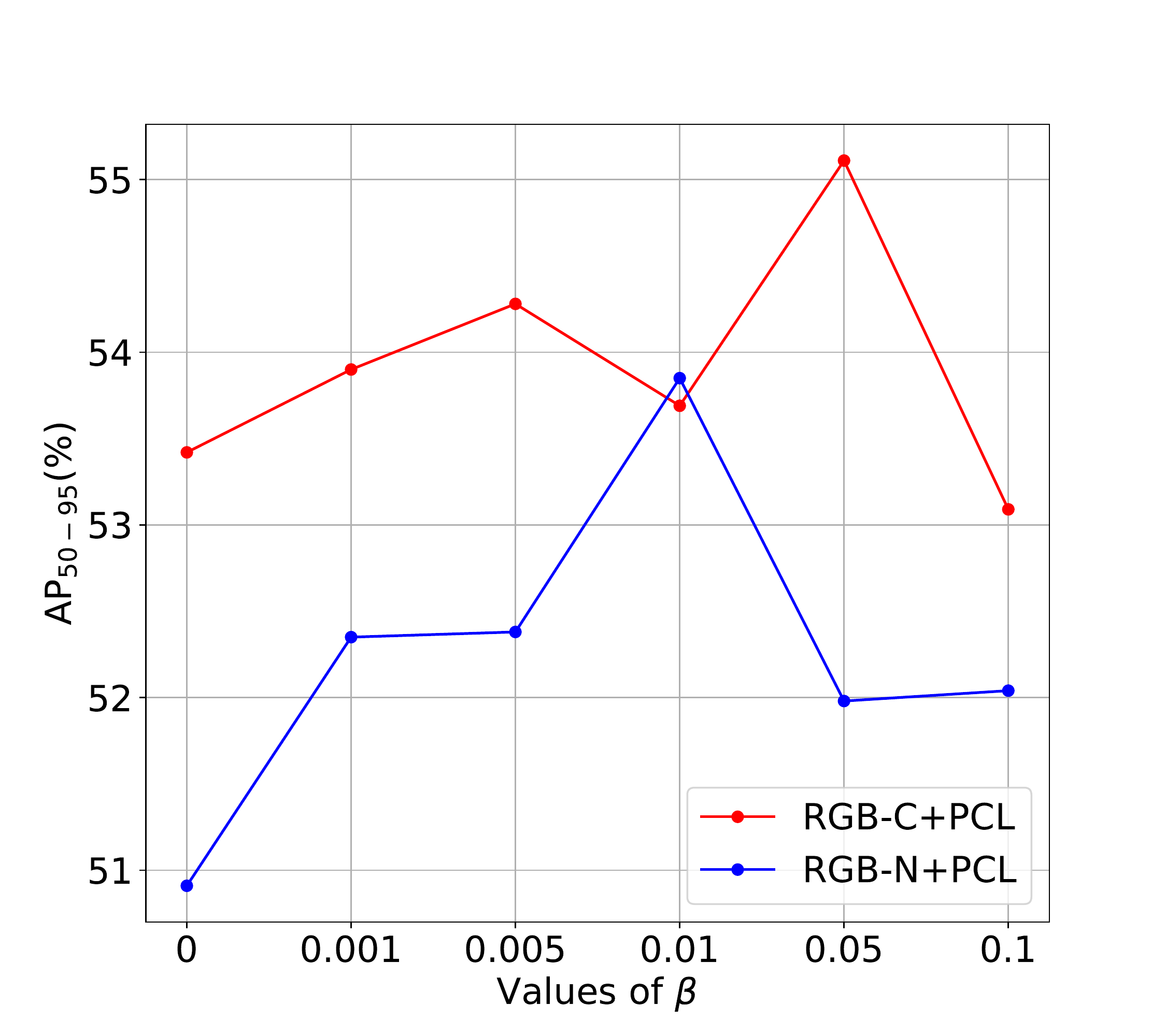}
\caption{Performance on COCO by varying values of $\beta$. It can be observed that large PCL weight would hurt the detection performance.}
\label{fig:sensity of beta}
\end{figure}

In Equ.~\eqref{equ: total}, we introduce a hyper-parameter $\beta$ to balance the PCL loss with the original RPN and RCNN loss. Here, we analyze the influence of the value of $\beta$. We vary the value of $\beta$ to [0.001, 0.005, 0.01, 0.05, 0.1], and the results in term of AP$_{50-95}$ are reported in Fig.~\ref{fig:sensity of beta}. It can be observed that the performance begins to drop when the weight of PCL loss is larger than 0.01 for RGB-N and 0.05 for RGB-C, which is mainly due to that too large weight will cause the model to be dominated by the auxiliary task and deviate from the main detection task. Thus, in default setting, the value of $\beta$ is set to 0.01 for RGB-N and 0.05 for RGB-C.

\subsubsection{Effectiveness of PCL for Unlabeled Data}
Here, we further conduct experiments to evaluate the effectiveness of PCL on unlabeled data. We randomly split the training data in COCO synthetic dataset to labeled data $D_l$ (50\%) and unlabeled data $D_u$ (50\%). In Table~\ref{tab: semi supervised}, Labeled PCL indicates applying PCL on the labeled data, Unlabeled PCL with RPN Score and Unlabeled PCL with RCNN Score mean applying PCL on unlabeled data with two different proposal sampling strategies as discussed in Sec.~\ref{sec: semi-super pcl}. The baseline here is RGB-N \cite{zhou2018learning} trained with labeled data $D_l$ which achieves 70.0\% of AP$_{50}$ and 44.6\% of AP$_{50-95}$. When PCL is applied on labeled data, we can obtain 71.0\% of AP$_{50}$ and 45.4\% of AP$_{50-95}$. When incorporating with unlabeled data $D_u$, we observe that using RCNN scores as guidance to construct sample pairs achieves better performance, which further enhances AP$_{50}$ to 73.0\% and AP$_{50-95}$ to 46.2\%. The overall improvement from unlabeled data is 3\% on AP$_{50}$ here. In Sec.~\ref{sec: experiments on unlabeled data}, we further investigate the effectiveness of PCL on unlabeled data for downstream standard datasets such as NIST16~\cite{guan2019mfc} and CASIA~\cite{dong2010casia,dong2013casia}, where we can obtain over 13\% improvement on AP$_{50}$. 

\begin{table}[t]
\centering
\caption{Experiments of PCL on unlabeled data with different proposal sampling strategies, RGB-N is adopted as the basic detector. $D_{l}$ indicates labeled data (50\% COCO training data), $D_{u}$ indicates unlabeled data (another 50\% COCO training data). Models are validated with three random labeled-unlabeled splits, the mean AP$_{50}$ (\%)/AP$_{50-95}$(\%) and standard variance are reported.}
  \label{tab: semi supervised}
  \begin{tabular}{cccc|cc}
    \toprule
      Training & Labeled & RPN & RCNN & \multirow{2}*{AP$_{50}$} & \multirow{2}*{AP$_{50-95}$} \\
      Data & PCL & \ Score \ & \ Score \ & & \\
     \midrule
      $D_{l}$ & $\times$ & $\times$ & $\times$ & 70.0$\pm$0.2 & 44.6$\pm$0.8  \\
      $D_{l}$ & \checkmark & $\times$ & $\times$ & 71.0$\pm$1.8 & 45.4$\pm$1.8 \\
      $D_{l}$ + $D_{u}$ & \checkmark & \checkmark & $\times$ & 72.5$\pm$1.4  & 45.5$\pm$1.1 \\
      $D_{l}$ + $D_{u}$ & \checkmark & $\times$ & \checkmark & \textbf{73.0$\pm$0.9} & \textbf{46.2$\pm$1.0} \\
  \bottomrule
\end{tabular}
\end{table}

\begin{table}[t]
\begin{center}
\caption{AP$_{50}$(\%) comparison on standard datasets.}
\label{tab:AP comparison with SOTA}
\begin{tabular}{cccccc}
\toprule
Methods  & NIST16 & Columbia & COVER & CASIA & IMD20\\
\midrule
RGB-N & 84.6 & \textbf{78.8} & 55.1 & 33.1 & 27.9  \\
+PCL(Ours) & \textbf{87.4} & 78.5 & \textbf{60.9} & \textbf{35.9} & \textbf{29.4} \\
\midrule
RGB-C & 82.2 & \textbf{81.3} & 62.0 & 39.3 & 31.4  \\
+PCL(Ours) & \textbf{86.2} & 79.8 & \textbf{62.7} & \textbf{39.3} & \textbf{32.5} \\
\bottomrule
\end{tabular}
\end{center}
\end{table}

\subsection{Experiments on Standard Datasets}
\label{sec: experiments on standard datasets}
In this section, we firstly evaluate our method on four standard image manipulation datasets NIST16~\cite{guan2019mfc}, CASIA~\cite{dong2010casia,dong2013casia}, COVERAGE~\cite{wen2016coverage},  Columbia~\cite{ng2009columbia} and one real-world dataset IMD20~\cite{novozamsky2020imd2020} on labeled data in supervised learning setting. Then we evaluate our methods on NIST16~\cite{guan2019mfc} and CASIA~\cite{dong2010casia,dong2013casia} datasets on unlabeled data in semi-supervised learning setting. We fine-tune the COCO pre-trained models on standard datasets except for the Columbia dataset which is directly tested on the COCO pre-trained models following previous work~\cite{zhou2018learning}. 

\subsubsection{PCL on Labeled Data}
We first conduct experiments in supervised learning regime with RGB-N and RGB-C as basic detectors respectively. We report AP$_{50}$ and AP$_{50-95}$ in Table~\ref{tab:AP comparison with SOTA} and Table~\ref{tab:AP 50-95 comparison with SOTA}. For both RGB-N and RGB-C, we find that PCL can consistently improve the AP among most datasets. The most significant improvement of AP$_{50}$ is on COVER for RGB-N, increased by 5.8\%, and on NIST16 for RGB-C, increased by 4\%. On Columbia, we observe our method achieves improvement in terms of AP$_{50-95}$, while shows degradation of AP$_{50}$. This means that the model misses more objects, while for the detected objects, the results tend to be more accurate. We speculate that this phenomenon possibly because the model is not fine-tuned on this dataset and the domain gap between datasets are the main reasons for the low recall. The consistent improvement shows that PCL can exploit the spatial relationship in tampered images and help to obtain more discriminative feature representations on labeled data. 

\begin{table}[t]
\begin{center}
\caption{AP$_{50-95}$(\%) comparison on standard datasets.}
\label{tab:AP 50-95 comparison with SOTA}
\begin{tabular}{cccccc}
\toprule
Methods & NIST16 & Columbia & COVER & CASIA & IMD20 \\
\midrule
RGB-N & 64.8 & 33.5 & 41.2 & 21.8 & 14.9 \\
+PCL(Ours) & \textbf{65.1} & \textbf{34.7} & \textbf{42.3} & \textbf{24.0} & \textbf{16.2} \\
\midrule
RGB-C & 61.1 & 35.8 & 40.2 & 25.8 & 17.5\\
+PCL(Ours) & \textbf{65.8} & \textbf{36.4} & \textbf{42.8} & \textbf{26.1} & \textbf{18.0}\\
\bottomrule
\end{tabular}
\end{center}
\end{table}

\begin{table}[t]
\begin{center}
\caption{AP$_{50-95}$(\%) comparison on different manipulation types on NIST16 dataset.}
\label{tab: different manipulation types}
\begin{tabular}{lcccc}
\toprule
Methods & Splicing & Copy-Move & Removal & Mean\\
\midrule
RGB-N &  \textbf{65.6} & 84.8 & 61.4 & 64.8 \\
+PCL(Ours) & 63.1 & \textbf{87.2} & \textbf{66.7} & \textbf{65.1} \\
\midrule
RGB-C & 62.0 & 70.2 & 49.1 & 61.1 \\
+PCL(Ours) & \textbf{62.8} & \textbf{82.5} & \textbf{64.9} & \textbf{65.8} \\
\bottomrule
\end{tabular}
\end{center}
\end{table}

Furthermore, we evaluate PCL with RGB-N and RGB-C as backbone on different manipulation types based on NIST2016 dataset. The results (in terms of AP$_{50-95}$) are listed in Table~\ref{tab: different manipulation types}. It is shown that PCL can obtain improvement on almost all the manipulation types. Most of the improvement stems from the success in removal. On RGB-N, we achieve 5.3\% AP$_{50-95}$ improvement on removal tampered images, while on RGB-C, we get 15.8\% improvement on removal tampered images.

\subsubsection{PCL on Unlabeled Data}
\label{sec: experiments on unlabeled data}
\begin{table*}[t]
\begin{center}
 \caption{Experiments on NIST16, CASIA and IMD20 under different ratios of labeled data in semi-supervised learning. AP$_{50}$ (\%) is reported.}
  \label{tab:sample ratio ap50}
  \begin{tabular}{llllllll}
    \toprule
    Datasets & Methods & 10\% & 30\% & 50\% & 70\% & 90\% \\
    \midrule
    \midrule
    \multirow{2}*{NIST16} & RGB-C & 29.1 & 45.5 & 64.2 & 68.0 & 76.7 \\
     & +PCL(Ours) & \textbf{65.8}\scriptsize{(+36.7)} & \textbf{67.0}\scriptsize{(+21.5)} & \textbf{77.8}\scriptsize{(+13.6)} & \textbf{80.4}\scriptsize{(+12.4)} & \textbf{86.4}\scriptsize{(+9.7)} \\
    \midrule
    \midrule
   \multirow{2}*{CASIA} &  RGB-C   & 28.4 & 30.0 & 33.2 & 35.5 & 38.1 \\
    & +PCL(Ours)  & \textbf{34.3}\scriptsize{(+5.9)} & \textbf{36.7}\scriptsize{(+6.7)} & \textbf{37.2}\scriptsize{(+4.0)} & \textbf{39.1}\scriptsize{(+3.6)} & \textbf{39.3}\scriptsize{(+1.2)} \\
    \midrule
    \midrule
    \multirow{2}*{IMD20} & RGB-C  & 15.5 & 21.2 & 23.9 & 29.8 & 30.6 \\
    & +PCL(Ours)  & \textbf{26.7}\scriptsize{(+11.2)} & \textbf{27.9}\scriptsize{(+6.7)} & \textbf{28.8}\scriptsize{(+4.9)} & \textbf{31.7}\scriptsize{(+1.9)} & \textbf{32.3}\scriptsize{(+1.7)} \\
  \bottomrule
\end{tabular}
\end{center}
\end{table*}

\begin{table*}[t]
\begin{center}
 \caption{Experiments on NIST16, CASIA and IMD20 under different ratios of labeled data in semi-supervised learning. AP$_{50-95}$ (\%) is reported.}
  \label{tab:sample ratio}
  \begin{tabular}{llllllll}
    \toprule
    Datasets & Methods & 10\% & 30\% & 50\% & 70\% & 90\%\\
    \midrule
    \midrule
    \multirow{2}*{NIST16} & RGB-C & 15.4 & 27.9 & 46.6 & 50.2 & 58.8\\
     & +PCL(Ours) & \textbf{31.2}\scriptsize{(+15.8)} & \textbf{38.9} \scriptsize{(+11.0)} & \textbf{51.0}\scriptsize{(+4.4)} & \textbf{56.4}\scriptsize{(+6.2)} & \textbf{63.3}\scriptsize{(+4.5)}  \\
    \midrule
    \midrule
    \multirow{2}*{CASIA} & RGB-C & 16.9 & 17.8 & 22.3 & 23.7 & 25.9\\
     & +PCL(Ours)  & \textbf{19.3}\scriptsize{(+2.4)} & \textbf{23.0}\scriptsize{(+5.2)} & \textbf{23.9}\scriptsize{(+1.6)} & \textbf{25.7}\scriptsize{(+2.0)} & \textbf{26.0}\scriptsize{(+0.1)}\\
    \midrule
    \midrule
    \multirow{2}*{IMD20} & RGB-C  & 7.2 & 11.1 & 13.0 & 16.4 & 16.2 \\
    & +PCL(Ours)  & \textbf{12.3}\scriptsize{(+5.1)} & \textbf{14.1}\scriptsize{(+3.0)} & \textbf{15.7}\scriptsize{(+2.7)} & \textbf{17.3}\scriptsize{(+0.9)} & \textbf{17.9}\scriptsize{(+1.7)} \\
  \bottomrule
\end{tabular}
\end{center}
\end{table*}

To demonstrate that PCL can be effectively used on real datasets with unlabeled data, we examine two semi-supervised learning settings. \textbf{First}, {\color{black} we randomly sample 10\%, 30\%, 50\%, 70\% and 90\% training data in NIST16, CASIA and IMD20 as the labeled set $D_{l}$ and regard the rest of training data as unlabeled set $D_{u}$.} Models are trained by the loss function Equ.~\eqref{equ: semi loss} with unlabeled data incorporated. \textbf{Second}, we use the entire training data as the labeled set and resort to the development dataset of NC2017 (the licensed dataset of OpenMFC 2017~\cite{guan2019mfc,guan2021user}) as the unlabeled set. The unlabeled set here is not necessarily in-domain. NC2017 has 2,512 images with different types of manipulation. 
Specifically, we combine the unlabeled images from NC2017 with COCO synthetic dataset for model pre-training. Then, we fine-tune the pre-trained models on CASIA, NIST16 and IMD20 datasets. As demonstrated in previous sections, RGB-C obtains better performance than RGB-N, we conduct experiments with RGB-C as the basic detector in the following sections. 

The AP$_{50}$ comparison is shown in Table~\ref{tab:sample ratio ap50}, and AP$_{50-95}$ comparison is shown Tabel~\ref{tab:sample ratio}. Since semi-supervised learning has not been widely studied for image manipulation detection task yet, we mainly compare PCL in semi-supervised regime with the vanilla supervised learning regime which only uses labeled data. On the three datasets NIST16 CASIA and IMB20, our method can consistently enhance the performance under different ratios of labeled data. We find PCL to be particularly effective under the setting of small ratios of labeled data, e.g. we get the improvement of 36.7\% AP$_{50}$ for NIST16 with 10\% labeled data (40 labeled images) and 6.7\% improvement for CASIA with 30\% labeled data (1533 labeled images). Interestingly, PCL is proven to be more data-efficient in semi-supervised settings than supervised learning for image manipulation detection task (65.8\% for PCL semi-supervised learning with 10\% labeled data v.s. 64.2\% for supervised learning with 50\% labeled data on NIST16). 
Furthermore, during training, we find that the model's performance is beneficial from longer training when more unlabeled data are incorporated which indicates that semi-supervised learning can reduce over-fitting in the low-label regime. 
From these results, we prove that PCL is effective for unlabeled data too. It is expected that we can obtain more generalizable feature representations with more data incorporated.

We summarize the experimental results in Table~\ref{tab:extra unlabeled data}, when the model is trained with 100\% labeled data and extra unlabeled data. It is shown that AP$_{50}$ on NIST16 can be improved by 7.1\%, AP$_{50}$ on CASIA is increased by 2.5\% and AP$_{50}$ on IMD20 is enhanced by 1.4\%. In this setting, we achieve state-of-the-art detection results on NIST16 (AP$_{50}$ of 89.26\%) and CASIA (AP$_{50}$ of 41.82\%). Here, we demonstrate that incorporating more extra realistic data in pre-training can benefit the downstream image manipulation detection.

\begin{table}[t]
\centering
\caption{Experiments on NIST16, CASIA and IMD20 in semi-supervised learning with 100\% labeled data and extra out-of-domain unlabeled data. AP$_{50}$ (\%) and AP$_{50-95}$ (\%) are reported.}
\label{tab:extra unlabeled data}
\begin{tabular}{lcccc}
\toprule
Methods & Training Data & NIST16 & CASIA & IMD20\\
\midrule
RGB-C & $D_{l}$ & 82.2/61.1 & 39.3/25.8 & 31.4/\textbf{17.5} \\
+PCL(Ours) & $D_{l} + D_{u}$ & \textbf{89.3/68.1} & \textbf{41.8/27.2} & \textbf{32.8/17.5}\\
\bottomrule
\end{tabular}
\end{table}

\begin{table}[t]
\centering
\caption{Ablation study of PCL in semi-supervised learning. $D_{l}$ indicates labeled data (50\% training data), $D_{u}$ indicates unlabeled data (another 50\% training data). Models are validated with three random labeled-unlabeled splits, the mean AP$_{50}$(\%) and standard variance are reported.}
\label{tab:semi-supervised}
\begin{tabular}{lccc}
\toprule
Methods & Training Data & NIST16 & CASIA\\
\midrule
RGB-C & $D_{l}$ & 64.2$\pm$2.1 & 33.2$\pm$0.8 \\
+PCL(Ours) & $D_{l}$ & 67.3$\pm$1.3  & 33.5$\pm$0.1 \\
+PCL(Ours) & $D_{l} + D_{u}$ & \textbf{77.8$\pm$0.9} & \textbf{37.2$\pm$0.5}\\
\bottomrule
\end{tabular}
\end{table}

\textbf{Ablation study of PCL in semi-supervised learning:} We further examine the effectiveness of PCL on labeled data and unlabeled data in semi-supervised learning. CASIA and NIST16 are split to 50\% labeled data $D_{l}$ and 50\% unlabeled data $D_{u}$. We evaluate the methods over three different data folds and report results in Table \ref{tab:semi-supervised}. It can be observed that when applying PCL on labeled data, we can improve AP$_{50}$ from 64.2\% to 67.3\% on NIST16, while on CASIA AP$_{50}$ is enhanced from 33.2\% to 33.5\%. When we further incorporate unlabeled data into training, more significant improvement is obtained (10.5\% on NIST16 and 3.7\% on CASIA), which shows that PCL is effective for both unlabeled data and labeled data.

\begin{table}[t]
\centering
\caption{Different choices of the threshold of RCNN scores $\delta$ in semi-supervised learning on NIST16. }
\label{tab:different choice of delta}
\begin{tabular}{ccc}
\toprule
$\delta$ & AP$_{50}$ (\%) & AP$_{50-95}$ (\%) \\
\midrule
0.4 & 75.2 & 51.0 \\
0.5 & 77.8 & 51.0\\
0.6 & \textbf{79.6} & \textbf{53.5}\\
0.7 & 75.6 & 48.8\\
\bottomrule
\end{tabular}
\end{table}

{\color{black} 
\textbf{Different choices of threshold of RCNN score $\delta$ in semi-supervised learning:} 
Different choices of $\delta$ indicate the correctness of the pseudo-label of proposals in unlabeled data. The larger values of $\delta$, the higher the correctness usually, however less number of proposals can be applied to PCL loss. We experiment with different values of $\delta$ on NIST16 dataset with 50\% labeled data and 50\% unlabeled data, the experimental results are shown in Table~\ref{tab:different choice of delta}. It can be observed that with different values of $\delta$, we can obtain better performance than only learning with 50\% labeled data (64.2\% AP$_{50}$ / 46.6\% AP$_{50-95}$), \textit{i.e.}, the method performs well under a wide range of hyper-parameters.
}

\subsubsection{Evaluation on Image-Level Manipulation Detection (Recognition)}
In this section, we demonstrate the effectiveness of PCL on image-level manipulation detection (recognition) task.  Image-level manipulation detection (recognition) aims to distinguish manipulated images from pristine ones via image-level binary classification, the evaluation also involves pristine images and reflects the false alarms of the models. Since CASIA are the only test dataset that owing the corresponding pristine images. We construct the test set with CASIA consisting of 50\% manipulated and 50\% pristine images as \cite{liu2022pscc}, noted as CASIA-D. We use {\color{black} image-level AUC and F1} for comparison. The experimental results are presented in Table~\ref{tab:image-level detection}. As our method is based on Faster-RCNN and makes no direct attempt to perform image-level recognition, we regard the max score among the detected boxes as the image-level binary classification score. For the compared methods, ManTraNet~\cite{wu2019mantra}, SPAN~\cite{hu2020span}, and PSCCNet~\cite{liu2022pscc} are segmentation-framework-based methods, the averages of the predicted maps are regarded as their scores.

As shown in Table~\ref{tab:image-level detection}, when RGB-C pre-trained on coco synthetic datasets with 11K data, we obtained the AUC of 72.04\%, when combined with PCL, the AUC can be further improved to 73.96\% which demonstrates that the proposed module PCL helps yield more discriminative features. As the compared counterparts trained their models with a large amount of data, we further pre-trained our models on a larger synthetic datasets released by PSCCNet~\cite{liu2022pscc} with 100K data. It is shown that RGB-C trained with 100K data achieved the AUC of 75.47\% and RGB-C+PCL outperforms with AUC of 76.38\% and F1 of 72.85\%.  

\begin{table}[t]
\centering
\caption{Image-level Manipulation Detection Evaluation on CASIA-D. Image-level AUC(\%) and F1(\%) are reported.}
\label{tab:image-level detection}
\begin{tabular}{lccc}
\toprule
Methods &  \# Data & AUC & F1\\
\midrule
ManTra-Net~\cite{wu2019mantra} & 64K & 59.94 & 56.69 \\ 
SPAN~\cite{hu2020span} & 96K & 67.33 & 63.48 \\
PSCC-Net~\cite{liu2022pscc} & 100K & 74.40 & 66.88 \\
\midrule
RGB-C & 11K & 72.04 & 71.08 \\
+PCL(Ours) & 11K & 73.96 & 71.18  \\
RGB-C & 100K & 75.47 & 72.31 \\
+PCL(Ours) & 100K & \textbf{76.38} & \textbf{72.85}  \\
\bottomrule
\end{tabular}
\end{table}

\begin{figure*}[t]
\centering
\includegraphics[width=0.99\textwidth]{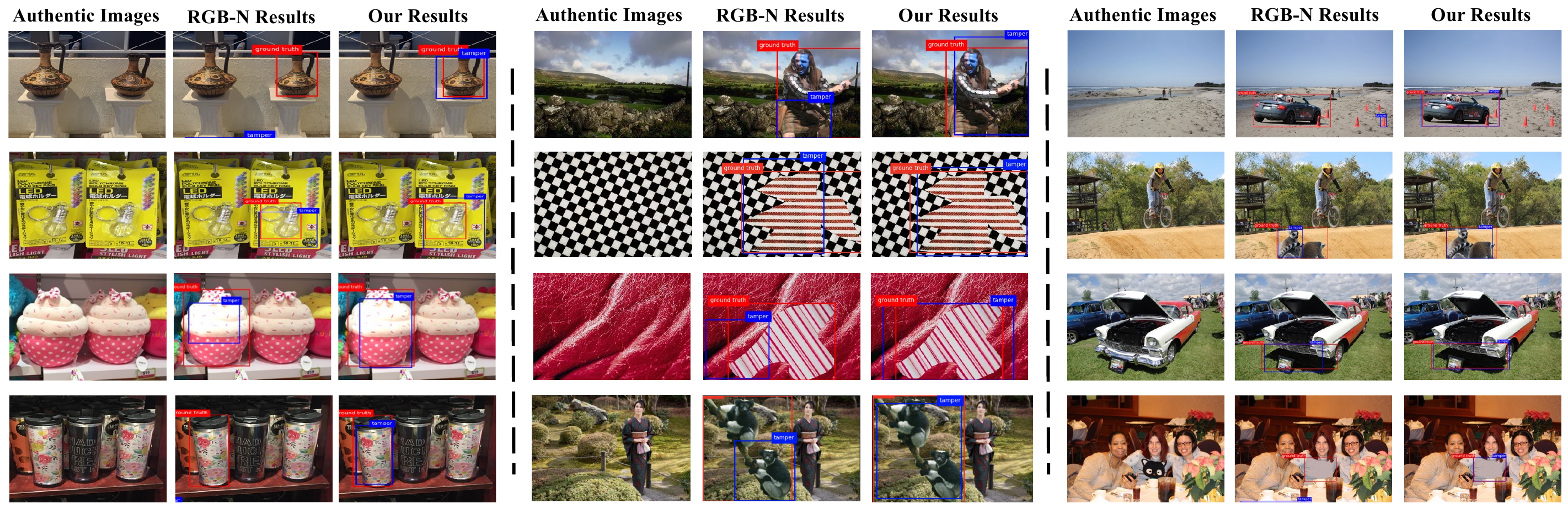}
\caption{Qualitative comparison between RGB-N and RGB-N+PCL. Mega columns from left to right are results from COVERAGE, CASIA V1 and NIST16. Red boxes are the ground-truth bboxes, blue ones are the predicted bboxes with highest scores. }
\label{fig:casia resutls}
\end{figure*}

\subsubsection{Comparison with Semantic-Segmentation-based Methods}
In the literature on image manipulation detection, another group is segmentation-based methods, such as classic unsupervised methods: ELA~\cite{ela2007}, NOI1~\cite{mahdian2009using}, CFA1~\cite{ferrara2012image} and deep learning-based methods: MFCN~\cite{salloum2018image}, SPAN~\cite{hu2020span}, GSR-Net~\cite{zhou2020generate}, MVSS-Net~\cite{chen2021image}, Mantra-Net~\cite{wu2019mantra} and PSCC-Net~\cite{liu2022pscc}. In this section, we compare our method with these models in terms of pixel-level metrics (F1 score and AUC score). To achieve this, we extend our model with a segmentation module as done in CR-CNN~\cite{yang2020constrained} and fine-tune the model end-to-end on downstream datasets. The results are reported in Table \ref{tab:f1 score} (the results of the compared counterparts are extracted from the corresponding papers).

\begin{table}[t]
\setlength\tabcolsep{4pt}
\begin{center}
\caption{Pixel-Level F1 and AUC score (\%) comparison with other counterparts on five standard datasets.}
\label{tab:f1 score}
\begin{tabular}{lccccc}
\toprule
Methods & NIST16 & Columbia & COVER & CASIA & IMD20 \\
\midrule
\multicolumn{5}{l}{\textit{classic unsupervised methods}}\\
ELA~\cite{ela2007} & 23.6/42.9 & 47.0/58.1 & 22.2/58.3 & 21.4/61.3 & -/-\\
NOI1~\cite{mahdian2009using} & 28.5/48.7 & 57.4/54.6 & 26.9/58.7 & 26.3/61.2 & -/- \\
CFA1~\cite{ferrara2012image} & 17.4/50.1 & 46.7/72.0 & 19.0/48.5 & 20.7/52.2 & -/- \vspace{0.4mm}\\
\multicolumn{5}{l}{\textit{semantic segmentation-based methods}}\\
MFCN~\cite{salloum2018image} & 57.1/- & 61.2/- & -/- & 54.1/- & -/- \\
SPAN~\cite{hu2020span} & 58.2/96.1 & \textbf{81.5}/\textbf{93.6} & 55.8/93.7 & 38.2/83.8 & -/75.0 \\
GSR-Net~\cite{zhou2020generate} & -/- & -/- & 48.9/- & \textbf{57.4}/- & -/-\vspace{0.4mm}\\
MVSS-Net~\cite{chen2021image} & 81.4/94.2 & -/- & 50.4/84.9 & 52.2/\textbf{87.7} & -/-\\
MantraNet~\cite{wu2019mantra} & -/79.5 & -/82.4 & -/81.9 & -/81.7 & -/74.8 \\
PSCC-Net~\cite{liu2022pscc} & 74.2/\textbf{99.1} & -/- & \textbf{72.3}/\textbf{94.1} & 55.4/87.5 & -/80.6 \\   
\multicolumn{5}{l}{\textit{object detection-based methods}}\\
RGB-N~\cite{zhou2018learning} & 72.2/93.7 & 69.7/85.8 & 43.7/81.7 & 40.8/79.5 & -/-\\
RGB-C(Ours) & \textbf{81.6}/95.6 & 67.0/74.4 & 61.3/88.3 & 46.3/74.9 & 49.9/\textbf{82.4}\\
+PCL(Ours) & 78.0/94.6 & 69.8/76.1 & 62.0/91.7 &  46.7/75.1 & 
\textbf{50.3}/82.3\\
\bottomrule
\end{tabular}
\end{center}
\end{table}

It can be observed that our methods can get better performance compared to classic unsupervised methods such as ELA~\cite{ela2007}, NOI~\cite{mahdian2009using}, CFA1~\cite{ferrara2012image} and object detection-based methods such as RGB-N~\cite{zhou2018learning}. While compared to semantic segmentation-based methods, our methods can outperform MFCN~\cite{salloum2018image}, SPAN~\cite{hu2020span}, GSR-Net~\cite{zhou2020generate}, MVSS-Net~\cite{chen2021image} and MantraNet~\cite{wu2019mantra} on NIST16, COVER and IMD20 datasets under the evaluation of F1, however, slightly lag behind SPAN~\cite{hu2020span} under the evaluation of AUC, which indicates that we can get higher precision, however the false positive rate is also higher (false positive predictions increase, meanwhile, true positive predictions grow more). We think that the reason for this phenomenon is that our approach is based on the detection framework, which makes more background areas in the bounding boxes have high responses. Most of semantic segmentation-based methods conduct pixle-wise prediction and design fine-gained segmentation modules, for example, both MFCN~\cite{salloum2018image} and GSR-Net~\cite{zhou2020generate} holds an edge prediction module which is especially beneficial for CASIA dataset (which contains lots of fine-grained manipulation), our approach lags behind them on CASIA. As SPAN adopts the pre-trained model released by~\cite{wu2019mantra}, which was pre-trained with millions of unreleased manipulation images, it outperforms others on Columbia. When compared with PSCC-Net~\cite{liu2022pscc}, we outperform on NIST16 (under the evaluation of F1) and IMD20 dataset. We lag behind on COVER and CASIA for the reason that our models are pretrained on an 11K COCO synthetic dataset following previous work~\cite{zhou2018learning}, while PSCC-Net is pretrained on a 100K synthetic manipulation dataset which contains various types of manipulation such as splice, copy-move and removal, thus PSCC-Net has a better generalization ability.

It is worth noting that our work is based on the object detection framework and is primarily designed to output the predicted bboxes, it has marginal advantages in terms of pixel-level evaluation metrics compared to the segmentation-based methods. However, in many real applications, there is actually no need to discover every tampered pixel. Instead, the complete and accurate bounding boxes are enough. More importantly, the semantic-segmentation-based methods must require per-pixel annotations for training, while the object-detection-based methods can be trained with the bounding-box annotations, which are much easier and cheaper than the per-pixel annotations (35$\times$ -- 85$\times$ improvement in annotation speed~\cite{kulharia2020box2seg}).
Therefore, the object-detection-based methods are also valuable, which can meet the need and do not require costly annotations, though they have a natural disadvantage in terms of pixel-level evaluation metrics.

\begin{figure}[t]
\centering
\includegraphics[width=0.35\textwidth]{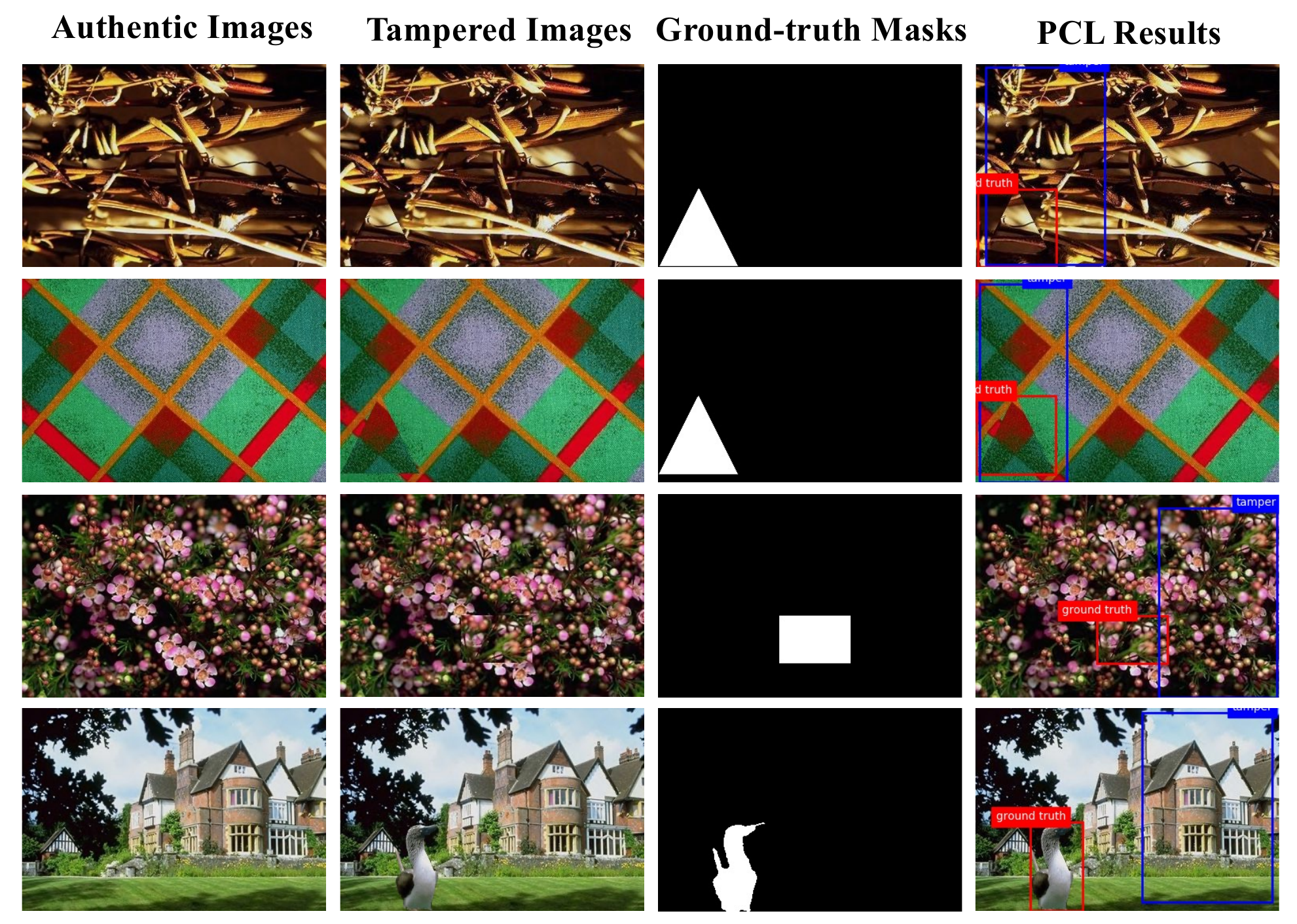}
\caption{Failure cases of RGB-N+PCL. Columns from left to right are: authentic images, tampered images, the masks of ground-truth and our detection results (red boxes indicates the ground-truth bounding boxes and the blue ones are our detection results).}
\label{fig:error analysis}
\end{figure}

\subsection{Qualitative Results}
\label{sec: qualitative results}
We show some qualitative results from COVERAGE, CASIA V1 and NIST16 in Fig.~\ref{fig:casia resutls}. We compare the detection results of RGB-N and RGB-N+PCL. The images displayed show that the proposal contrastive learning can achieve better performance in both manipulation classification and localization. 

\textbf{Error Analysis.} We also provide some error analysis of our method in Fig.~\ref{fig:error analysis}. We find that most of the failure cases of our method are indeed hard examples, especially the images with manipulated regions that are very similar to the background. The failure cases show that although PCL has improved the performance of manipulation detection, it still cannot identify some cunning tampering well, which may require more specific methods. 

\section{Limitation and Future Work}

The proposed Proposal Contrastive Learning (PCL) method facilitates the feature representation in manipulation detection models and helps to make full use of extra unlabeled data which we believe can offer insights for practical applications. Though we demonstrate the advantage of our method with the consistent improvement in AP in both supervised learning and semi-supervised learning settings, our method has marginal advantages in terms of pixel-level evaluation metrics compared to the segmentation-based methods. 

In future work, (i) we plan to extend the local contrastive learning idea to the segmentation-based methods to enhance the pixel-level localization performance; (ii) further explore semi-supervised learning in the area of image manipulation localization; (iii) moreover, large-scale pre-training data can be utilized to improve the ability of image manipulation models in dealing with the more challenging real-life manipulated images.

\section{Conclusion}
In this paper, we propose a Proposal Contrastive Learning (PCL) module for image manipulation detection. Our method is deployed on a two-stream architecture by contrasting the features from different views (e.g., RGB and noise features here). PCL can capture the view-invariant factors between different views and thus enhance the feature representations. Moreover, our method can be adapted to unlabeled data to alleviate the problem of insufficient training samples. Extensive experiments among several standard datasets show the effectiveness of PCL for image manipulation detection.

\bibliographystyle{IEEEtran}
\bibliography{ref}

\end{document}